\documentclass[letterpaper, 10pt, conference]{ieeeconf}

\IEEEoverridecommandlockouts
\usepackage[utf8]{inputenc}
\usepackage[T1]{fontenc}
\usepackage{amsmath,amsfonts}
\usepackage{tabularx}
\usepackage{algorithmic}
\usepackage{algorithm}
\usepackage{array}
\usepackage[caption=false,font=normalsize,labelfont=sf,textfont=sf]{subfig}
\usepackage{textcomp}
\usepackage{stfloats}
\usepackage{url}
\makeatletter
\g@addto@macro{\UrlBreaks}{\UrlOrds}
\makeatother
\usepackage{verbatim}
\usepackage{graphicx}
\usepackage{tikz}
\usepackage{caption}  
\usepackage{cuted}      
\usepackage{float}
\usepackage{cite}
\usepackage[hidelinks]{hyperref}  
\usepackage{booktabs}               
\usepackage{threeparttable}
\usepackage{multirow}               
\usepackage{makecell}               
\usepackage{pifont}                 
\usepackage{xcolor}
\newcommand{\cmark}{{\color{green!80!black}\ding{51}}}      
\newcommand{\xmark}{{\color{red!80!black}\ding{55}}}        
\newcommand{\meh}{$ \color{orange!170!yellow} \boldsymbol{\sim}$ }

\usepackage[most]{tcolorbox}
\newif\ifuseRevisions

\newcommand{\rev}[1]{\ifuseRevisions\textcolor{blue}{#1}\else #1\fi}

\begin{document}

\title{\LARGE \bf SwarmNxt: Open-source Software-Hardware Platform for Fast and Agile Aerial Swarms}

%

\author{Charbel Toumieh, Niel Mistry, Benjamin Jarvis, \\ Simon Jeger, Peize Liu, Shaojie Shen, Dario Floreano,~\IEEEmembership{Fellow,~IEEE}
\thanks{C. Toumieh, N. Mistry, B. Jarvis, S. Jeger, and D. Floreano are with the Laboratory of Intelligent Systems, Ecole Polytechnique Federale de Lausanne (EPFL), CH1015 Lausanne, Switzerland.}%
\thanks{P. Liu and S. Shen are with the Department of Electronic and Computer Engineering, Hong Kong University of Science and Technology, Hong Kong.}%
\thanks{This work was supported by the Swiss National Science Foundation (SNSF) with grant number 200020\textunderscore212077, and Armasuisse grant number 591797.}%
\thanks{\raggedright \rev{Video: \url{https://youtu.be/9aOr5EDLQEo}}}%
\thanks{\rev{Code and documentation: \url{https://github.com/lis-epfl/swarm-nxt}}}%
}

\bstctlcite{BSTcontrol}
\maketitle
\thispagestyle{empty}
\pagestyle{empty}

\begin{strip}
\vspace{-70pt}  
\centering
  \begin{tikzpicture}
      \node[anchor=south west, inner sep=0] (image) at (0,0) {
          \includegraphics[width=\textwidth]{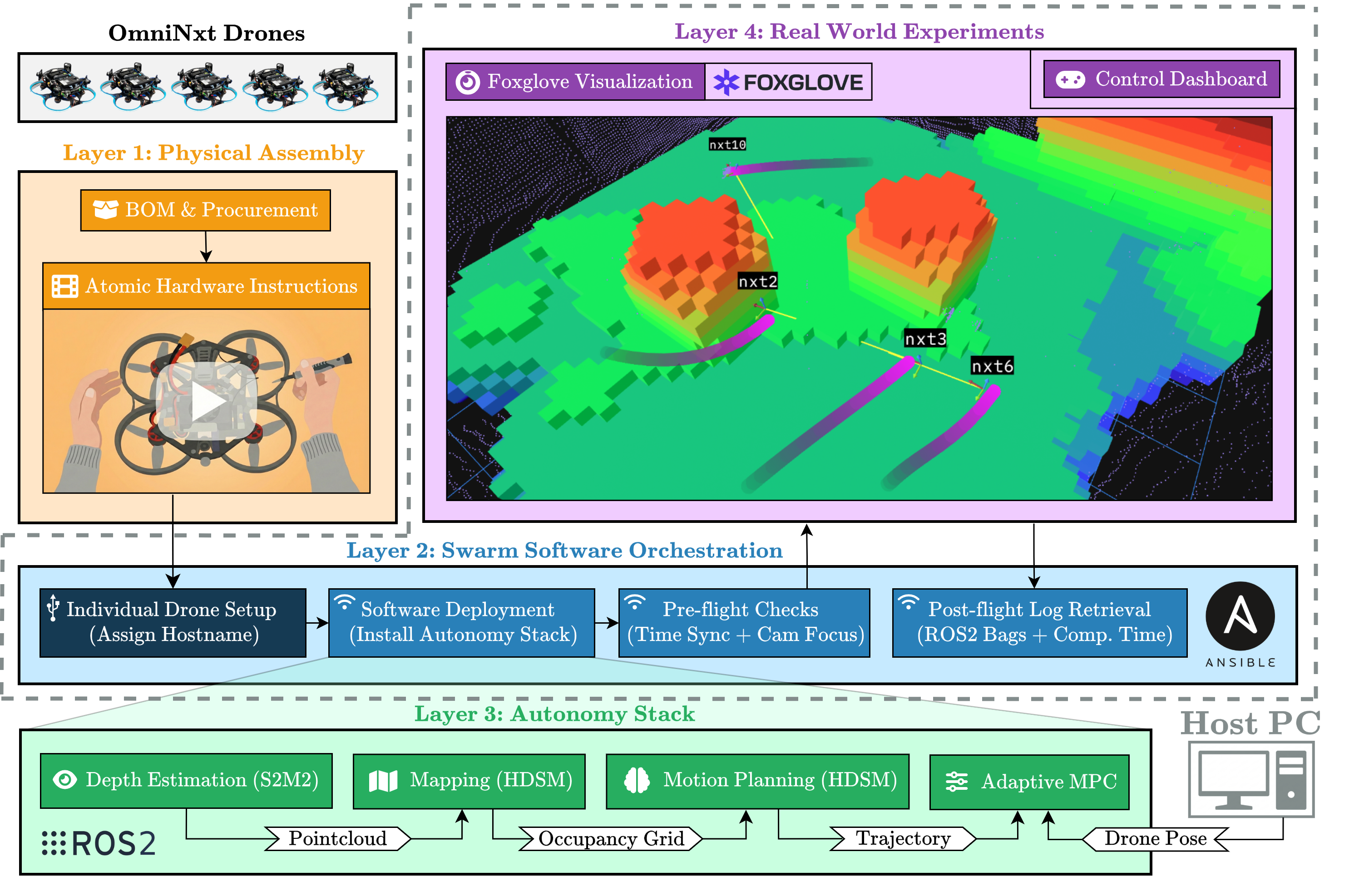}
      };
  \end{tikzpicture}
  
  \captionof{figure}{
    Overview of the SwarmNxt framework. The pipeline spans four layers: (1) physical assembly with a bill of materials and atomic build instructions, (2) Ansible-based swarm software orchestration for parallel deployment and pre/post-flight validation, (3) a ROS 2 autonomy stack integrating depth estimation (S2M2), mapping and planning (HDSM), and adaptive MPC, and (4) real-world experiment monitoring via Foxglove and a control dashboard.
  }
  \label{fig:overview}
\end{strip}
 \begin{abstract}
   Aerial robot swarms have the potential to transform time-critical safety, security, and search-and-rescue operations. By coordinating multiple robots, they can rapidly survey disaster sites, map collapsed or GPS-denied environments, and search cluttered areas faster than a single robot, reducing response times and minimizing risks to first responders. Realizing this potential, however, requires robust autonomous swarm navigation, which remains an active research challenge. Progress is further constrained by existing platforms, as commercial drones are often closed-source or lack the onboard computational resources needed for agile, vision-based collective flight. Moreover, developing, deploying, and maintaining software across multiple aerial robots requires significant engineering effort.
To address these challenges, we present SwarmNxt, an open-source software platform built on the open-source OmniNxt drone hardware. SwarmNxt provides an end-to-end toolkit, including detailed hardware assembly instructions with a video tutorial, automation tools for parallel software deployment and swarm-wide updates, and a ROS 2–based framework for autonomous navigation. The platform integrates state-of-the-art control, planning, and depth estimation \rev{into a single ROS~2 multi-agent system, providing an open research infrastructure for physical swarm experimentation}.
We validate SwarmNxt through two real-world experiments: a six-drone swarm performing decentralized planning with high-speed inter-drone collision avoidance, and a four-drone swarm executing collective flight with onboard depth estimation in an obstacle-filled environment. \rev{Both experiments were run indoors with global position from external motion capture; perception, planning, and control run onboard.}
\end{abstract}


\section{INTRODUCTION}
Aerial robot swarms promise to improve time-critical safety, security, and search-and-rescue missions. By distributing tasks across multiple aerial robots, a swarm can rapidly explore disaster sites, inspect collapsed or GPS-denied structures, and search complex environments more efficiently than a single drone, while reducing the exposure of first responders to hazardous conditions. \rev{For the search-and-rescue and civil-protection teams who would field them (the end users), such systems must additionally tolerate the loss of individual units and be operable by personnel who are not roboticists.} Despite this potential, enabling fully autonomous swarm operation in such scenarios remains an active area of research. Reliable navigation in unknown environments, onboard perception and decision-making, and resilient decentralized coordination are still challenging problems that require further advances\rev{; recent overviews of aerial swarm robotics survey these open problems in detail~\cite{dorigo2026drones,chen2020towardrobust,du2025survey}}.

\rev{One of the main obstacles to advancing swarm autonomy is the absence of a platform with the compute and sensing required for advanced autonomy algorithms, as well as for fast and safe experimentation: one combining omnidirectional perception, onboard compute for learned depth estimation alongside planning and control, ROS~2 multi-agent scalability, fleet-level deployment tooling, and reproducible documentation (Table~\ref{tab:platform_comparison}).} Existing commercial drones are frequently closed-source or provide insufficient onboard computational capability for algorithms involving real-time perception, planning, and inter-robot coordination. In addition, scaling experiments from a single robot to a swarm introduces significant software engineering challenges, including deployment, configuration, synchronization, and maintenance across multiple vehicles. These practical limitations increase the cost and complexity of experimental validation, slowing the development of autonomous swarm systems.

\begin{table*}[htbp]
\centering
\begin{threeparttable}
\caption{Platform Comparison. Size: Diagonal distance with propellers, Open: Open Source, Ext: Extensible, Swarming: algorithms and control tools for autonomous swarm navigation, Scripts: Automated setup/update scripts, Atomic: Atomic instructions, $\sim$: Partial.}
\label{tab:platform_comparison}
\small 
\setlength{\tabcolsep}{3.0pt} 
\renewcommand{\arraystretch}{1.3}
\begin{tabular}{l ccccccc cccc c} 
\toprule
& \multicolumn{7}{c}{\textbf{Hardware}} & \multicolumn{4}{c}{\textbf{Software}} & \textbf{Docs} \\
\cmidrule(lr){2-8} \cmidrule(lr){9-12} \cmidrule(lr){13-13}
\textbf{Platforms} & \textbf{Size (m)} & \textbf{Mass (g)} & \textbf{Open} & \textbf{Ext.} & \textbf{Perception} & \textbf{FOV} & \textbf{GPU} & \textbf{Open} & \textbf{Framework} & \textbf{Swarming} & \textbf{Scripts} & \textbf{Atomic} \\
\midrule

DJI Mavic 3E \cite{djiMavicEnterprise} & 0.62 & 920 & \xmark & \xmark & Fisheye & 360° & \xmark & \xmark & Proprietary & \xmark & \xmark & \xmark \\
Skydio X10 \cite{skydio_x10_specs} & $\approx$1 & 2490 & \xmark & \xmark & Fisheye & 360° & \cmark & \xmark & Proprietary & \xmark & \xmark & \xmark \\
FLA \cite{mohtaFastAutonomousFlight2018} & 0.7 & $\approx$2500 & \cmark & \cmark & Cam + LiDAR & $\approx$90° & \xmark & \cmark & ROS 1 & \xmark & \xmark & \xmark \\
Fast 250 \cite{zjufast2022fastdrone250} & 0.4 & $\approx$100 & \cmark & \cmark & Stereo Cam & $\approx$90° & \xmark & \cmark & ROS 1 & \xmark & \meh & \cmark \\
Agilicious \cite{foehnAgiliciousOpensourceOpenhardware2022} & 0.45 & $\approx$775 & \cmark & \xmark & Stereo Cam & $\approx$90° & \cmark & \cmark & ROS 1 & \xmark & \xmark & \xmark \\
MRS\textsuperscript{*} \cite{bacaMRSUAVSystem2021} \cite{hertBacaMRSHardware2022} & 0.7 & $\approx$1700 & \cmark & \cmark & Cam + LiDAR & $\approx$90° & \cmark & \cmark & ROS 1$\sim$2 & \cmark & \meh & \meh \\
Crazyflie \cite{crazyflie} \cite{crazyswarm}  & 0.092 & $\approx$27 & \cmark & \cmark & Mono Cam & $\approx$90° & \xmark & \cmark & ROS 2 & \cmark & \cmark & \cmark \\
Starling 2\textsuperscript{$\dagger$} \cite{modalai2024starling2} & 0.22 & 280 & \cmark & \cmark & Fisheye & 170° & \cmark & \cmark & ROS 2 & \xmark & \meh & \cmark \\

OmniNxt \cite{liuOmniNXT} & 0.27 & 660 & \cmark & \cmark & Fisheye & 360° & \cmark & \cmark & ROS 1 & \xmark & \meh & \meh \\

\midrule
\textbf{SwarmNxt (ours)} & 0.27 & 660 & \cmark & \cmark & Fisheye & 360° & \cmark & \cmark & \textbf{ROS 2} & \textbf{\cmark} & \cmark & \cmark \\
\bottomrule
\end{tabular}

\vspace{0.5em}

\begin{tablenotes}
    \small
    \item[*] Currently transitioning to ROS 2; multiple modules require months of work before completion per the project's GitHub \cite{mrs_uav_system_github}.
\vspace{0.3em}
    \item[$\dagger$] The GPU is Adreno 650 \cite{qualcomm2020adreno650}; other GPU platforms in the table use NVIDIA Orin NX 16GB \cite{nvidia2024jetsonorinxdatasheet} which provides 10x more compute.
\end{tablenotes}
\vspace{0.3em}
\end{threeparttable}
\end{table*}

\begin{table}[t]
\vspace{5pt}
\centering
\begin{threeparttable}
\caption{Temporal Overview of Swarm Deployment}
\label{tab:timing}
\small 
\renewcommand{\arraystretch}{1.3} 
\begin{tabular}{l l l}
\toprule
\textbf{Task} & \textbf{Frequency} & \textbf{Duration} \\
\midrule
BOM \& Procurement\tnote{*} & \textcolor{red}{One-time} & 1.5 months \\
Hardware Assembly & \textcolor{red}{One-time} & 5.0 hr (per drone) \\
\midrule
Individual Drone Setup & \textcolor{red}{One-time} & 1.5 hr (per drone) \\
Parallel Deployment\tnote{$\dagger$} & \textcolor{blue}{Recurring} & 20 min / 1 min \\
Pre-flight Validation & \textcolor{green!60!black}{Per Flight} & 20 seconds \\
Post-flight Log Retrieval & \textcolor{green!60!black}{Per Flight} & 30 seconds \\
\bottomrule
\end{tabular}

\vspace{0.5em}

\begin{tablenotes}
    \small
    \item[*] Duration based on shipping to Switzerland.
\vspace{0.3em}
    \item[$\dagger$] Duration for initial installation vs. recurring updates via Ansible.
\end{tablenotes}
\end{threeparttable}
\end{table}

A review of current aerial platforms (Table \ref{tab:platform_comparison}) reveals a gap between deployment simplicity and research-grade capabilities. Commercial solutions like the DJI Mavic 3E \cite{djiMavicEnterprise} and Skydio X10 \cite{skydio_x10_specs} offer reliability, 360° perception, \rev{regulatory certification, and turnkey use without in-house expertise}, but their closed-source nature prevents algorithmic customization and integration with cutting-edge research frameworks. Among open-source alternatives, nano-drone platforms like the Crazyflie \cite{crazyflie} excel at swarm demonstrations \cite{crazyswarm} \cite{pichierri2023crazychoir} with comprehensive automation scripts and atomic build instructions, yet they lack the computational power and perception necessary for autonomous navigation in complex environments. 

Conversely, more powerful research platforms provide robust capabilities but face critical deployment limitations. The FLA \cite{mohtaFastAutonomousFlight2018} and Fast 250 \cite{zjufast2022fastdrone250} pioneered open-source autonomous flight with ROS 1, while Agilicious \cite{foehnAgiliciousOpensourceOpenhardware2022} added GPU-accelerated processing. However, ROS 1's single-master architecture fundamentally limits scalability in multi-agent systems, and none of these platforms address the growing complexity of deploying and maintaining consistent software across multiple agents. While the Starling 2 \cite{modalai2024starling2} provides a modern ROS 2 foundation with atomic documentation, it requires researchers to develop their own swarm coordination frameworks and lacks the onboard compute to run real-time depth estimation alongside planning and control. 

The MRS system \cite{bacaMRSUAVSystem2021} \cite{hertBacaMRSHardware2022} represents the current state-of-the-art for swarm-capable platforms, offering both hardware extensibility and multi-agent coordination capabilities in ROS 1, with partial ROS 2 support and setup scripts currently under development. However, its transition to ROS 2 remains incomplete, requiring months of additional work per the project's documentation \cite{mrs_uav_system_github}, and atomic build instructions are limited.  Finally, OmniNxt \cite{liuOmniNXT} offers an excellent hardware foundation with 360° omnidirectional vision and GPU capabilities, but lacks a dedicated framework for swarm-wide orchestration, automated parallel deployment scripts, and ROS 2-based multi-agent coordination \cite{macenski2022ros2}.

To address these limitations, we introduce SwarmNxt (Figure \ref{fig:overview}), an open-source framework designed to lower the barrier of entry for swarm research by providing a turnkey solution that streamlines the transition from hardware assembly to multi-robot deployment. While the hardware described here is based on the OmniNxt drone \cite{liuOmniNXT}, the software orchestration tools and autonomy stack can be adapted to other drones using the NVIDIA Orin series \cite{nvidia2024jetsonorinxdatasheet} and PX4 \cite{px4_autopilot} with minimal effort. \rev{While the platform is built for indoor experimentation and does not transfer as-is to GPS-denied field environments, it is designed to make research on the modules such deployment depends on---control, planning, SLAM, and depth estimation---safe, fast, and iterative.} The primary contributions of this work are four-fold: 1) a comprehensive hardware pipeline featuring atomic build instructions and a video tutorial to minimize assembly overhead (Section \ref{sec:atomic_hardware}); 2) a suite of automation scripts for parallel software configuration and swarm-wide updates (Section \ref{sec:auto_scripts}); 3) an extensible ROS 2 framework \rev{that integrates existing state-of-the-art algorithms for control~\cite{Verschueren2021}, planning~\cite{toumiehHighSpeedPlanning}, and vision-based depth estimation~\cite{min2025s2m2}, together with the multi-agent adaptations required to operate them on a physical swarm} (Section \ref{sec:autonomy_stack}); and 4) experimental validation on physical swarms of six drones performing coordinated planning with recursive collision avoidance in an empty environment, and four drones utilizing onboard depth estimation in an environment with obstacles (Section \ref{sec:exp_validation}).

\section{Framework Architecture}
This section details the pipeline of the SwarmNxt framework (Figure \ref{fig:overview}), which enables rapid deployment of aerial drone swarms. An overview of the time required for each stage of the assembly and deployment process is provided in Table \ref{tab:timing}.

\subsection{Physical Assembly} \label{sec:atomic_hardware}
The initial technical hurdle in swarm research is the physical procurement and assembly of multiple identical agents.

We provide a comprehensive Bill of Materials (BOM) with direct vendor links\rev{, totalling approximately 2300~CHF per drone at the time of writing,} and atomic hardware build instructions paired with a video tutorial of the assembly. These granular resources allow researchers without specialized mechanical expertise to assemble an  OmniNxt drone in approximately five hours.

\begin{figure}
\centering
  \includegraphics[width=0.485\textwidth, trim={0.2cm 0cm 0 0.cm}, clip]{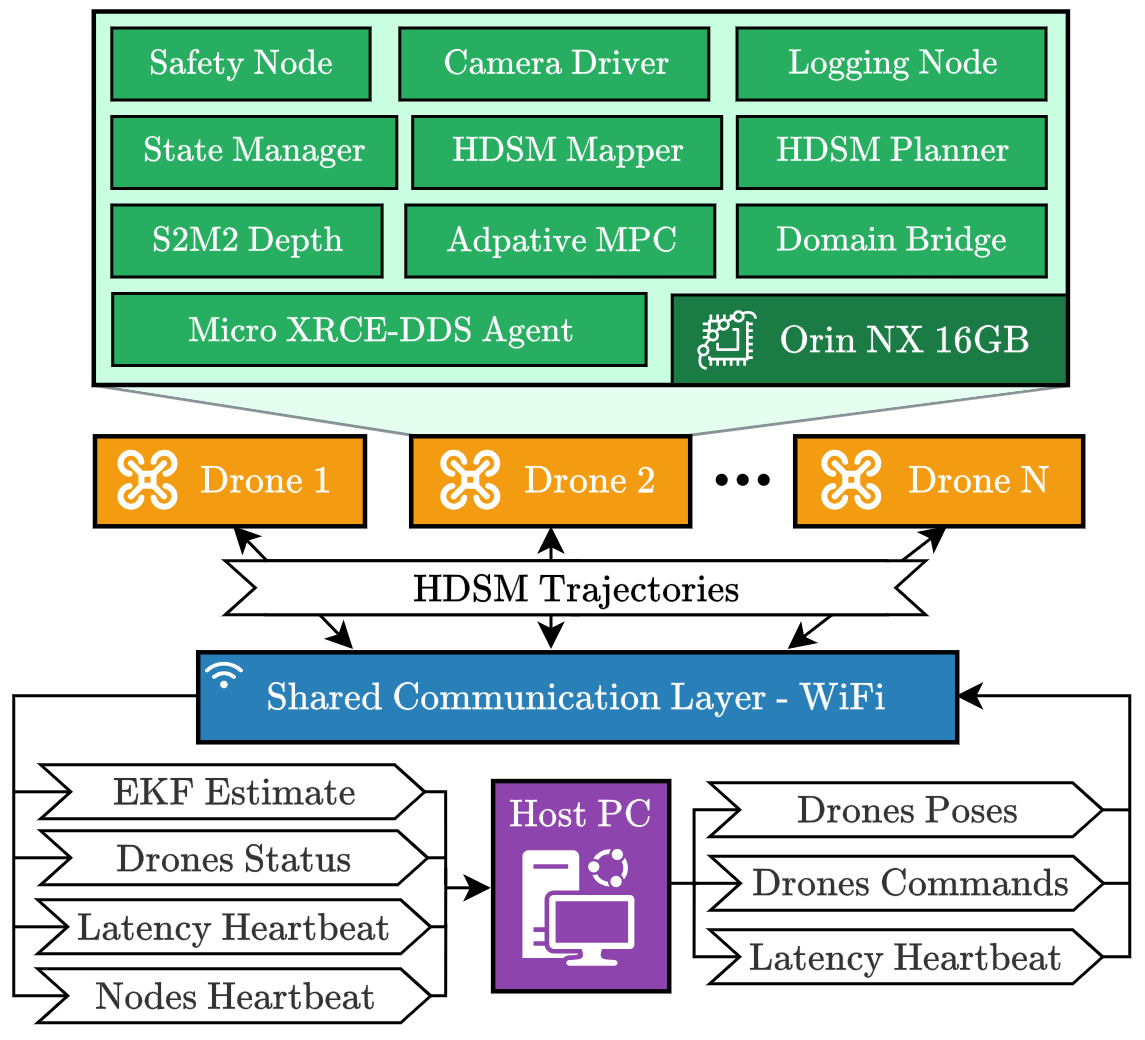}
  \caption{
    Overview of the communication architecture and the nodes running on each drone.
  }
  \label{fig:com_diagram}
\end{figure}

\subsection{Swarm Software Orchestration} \label{sec:auto_scripts}
Managing software consistency and operational readiness is a significant bottleneck as the number of drones increases. We solve this through Ansible-based automation scripts (called playbooks) \cite{ansible_automation_platform} that perform parallel configuration and validation across the entire fleet. \rev{Ansible is used unmodified as the execution engine; all four playbooks below were written for this work.}

\subsubsection{Individual Drone Setup} Following the base OS installation using the NVIDIA SDK \cite{nvidia_jetpack}, an Ansible playbook is executed on each drone individually. This playbook is essential for assigning a unique hostname to each unit, allowing the system to distinguish between agents and enabling all subsequent parallelization. It also handles system-level tasks such as configuring serial port access and optimizing power profiles for maximum computational performance.

\subsubsection{Parallel Software Deployment} Once hostnames are established, a secondary setup playbook is run in parallel across the entire swarm. This playbook automatically clones and builds the latest ROS 2 autonomy packages from GitHub and handles all dependency installations, ensuring every agent is running the latest software. If a package has been updated and needs to be pulled from GitHub again, the playbook can simply be rerun. Due to Ansible's idempotency, it automatically detects which packages have changed and updates/rebuilds only those, with minimal overhead. 

\subsubsection{Pre-flight Validation} A pre-flight playbook performs critical checks to ensure that onboard cameras are synchronized and focused, and that communication latencies and time synchronization between all the drones and the host PC are within acceptable bounds before takeoff.

\subsubsection{Post-flight Log Retrieval} Following flight operations, a playbook retrieves system logs and ROS-bag files from all drones simultaneously. This centralizes data on the Host PC for rapid multi-agent performance analysis and synchronized debugging. Additionally, it displays the MPC tracking error, computation times of the autonomy stack, and the minimum inter-agent distance during the flight for a quick evaluation of the flight performance. 

\subsection{ROS 2 Autonomy Stack} \label{sec:autonomy_stack}
The ROS 2 autonomy stack is built on a DDS-based communication layer \cite{eprosima_fastdds} that enables reliable, low-latency information exchange between the different autonomy modules (Figure \ref{fig:com_diagram}).
 To ensure the system remains organized as it scales, each drone operates on its own isolated network domain (ROS Domain ID). This mitigates broadcast storms—where agents are flooded with messages from all other drones—and reduces the computational load on the onboard Orin NX. This isolation is specifically critical for the Micro XRCE-DDS interface \cite{micro_xrce_dds}, which manages the high-speed link between the drone's computer and its PX4 flight controller; by keeping each drone on a separate domain, we protect vital flight commands from being delayed by external network traffic.

A centralized Host PC operates on a primary domain (ROS Domain ID 0) to oversee the entire swarm. We use a domain bridge running on each drone and the Host PC to allow only essential information to cross between these domains. For example, the bridge broadcasts each drone's planned trajectory to all other agents to support decentralized collision avoidance. \rev{Planning, mapping, and control themselves run onboard each agent, and trajectories are exchanged directly between drones, whereas ground-truth position is supplied by motion capture through the Host PC; we therefore use ``decentralized'' for the planning algorithm rather than for the system as a whole, which retains this central dependency.}

\subsubsection{Autonomous Navigation Stack} The framework integrates a state-of-the-art autonomous navigation stack
consisting of the High-speed Decentralized and Synchronous Motion planner
(HDSM) \cite{toumiehHighSpeedPlanning} for mapping and motion planning, an
adaptive Model Predictive Controller (MPC) \cite{Verschueren2021} for robust
trajectory tracking, and the Scalable Stereo Matching Model (S2M2)
\cite{min2025s2m2} for vision-based depth estimation. \rev{HDSM builds on earlier work on voxel-grid safe corridors~\cite{toumiehVoxelGrid2022,toumiehShapeAware2022}, decentralized planning with time-aware safe corridors~\cite{toumiehDecentralizedMultiAgentPlanning2022}, and latency-robust multi-agent planning~\cite{toumiehLatency2023}.} Global positioning is provided by an external motion capture system \cite{optitrack} and communicated to each agent via the Host PC.

The adaptive MPC tracks reference trajectories generated by the HDSM planner while accounting for the actuator limits and quadrotor dynamics at 100~Hz. The controller takes as input the drone's current state and the planned trajectory, and outputs collective thrust and body rate commands sent to the PX4 flight controller. To compensate for battery voltage drop during flight, the MPC output is scaled by a factor derived from the vertical tracking error. When the drone operates below the reference altitude, the scaling factor is increased; when above, it is decreased, following an integral control law. This simple adaptation significantly improves altitude stability over extended flight durations.

For depth estimation, we employ S2M2 \cite{min2025s2m2}, a learning-based stereo matching network. The model takes as input a pair of rectified stereo images and outputs a dense depth map by predicting per-pixel disparity. We use the small variant (26.5M parameters) with an input resolution of 256$\times$160, selected empirically to balance depth accuracy with real-time performance on the onboard compute. \rev{Larger variants were not viable: the network already occupies 95\% of the GPU (Figure~\ref{fig:comp_time}), so added capacity would lower the update rate rather than improve navigation.}
The four onboard fisheye cameras are calibrated via TartanCalib \cite{duisterhof2022tartancalib} and rectified to pinhole projections to form four virtual stereo pairs, providing omnidirectional depth coverage. Each pair is fed to S2M2 to produce a depth map, resulting in four depth maps per inference cycle. In quantitative evaluation, the estimated depth exhibited an average error of approximately 10~cm at a distance of 2~m. To ensure collision-free navigation, this uncertainty is accounted for by inflating all detected obstacles by an additional margin during mapping.

The HDSM mapping pipeline was modified to support multi-drone operation and noisy depth measurements. Since the map is intended to represent only static obstacles, point-cloud segments corresponding to other drones are removed prior to integration in the occupancy map. This filtering is performed by identifying points within a bounding box centered at the current position of each neighboring agent. To handle depth uncertainty, we extended the mapping formulation to use a log-likelihood occupancy update scheme \cite{thrun2005probabilistic}. Rather than assigning voxels discrete states, each observation increases the occupancy value of a voxel when measured as occupied, and decreases it when freed through computationally efficient raycasting \cite{toumieh2021gpu}. Voxels are subsequently classified as free if their value falls below -1, occupied if above 1, or unknown otherwise. This formulation improves map stability in the presence of noisy depth estimates while preserving real-time performance.

\subsubsection{Safety Supervision} \label{sec:safety} A Safety Node acts as a technical supervisor for each drone to maintain experimental integrity. It automatically triggers a landing command if an agent exits a predefined safe zone (e.g. the tracking area of the motion capture system) or if the EKF variance exceeds a critical threshold, indicating high positional uncertainty. In the event of a low battery, an emergency landing is automatically triggered by the flight controller; therefore, this case was not included in the safety node.

\subsubsection{Swarm Command and Visualization} To facilitate real-time interaction, we provide a unified drone control dashboard\rev{\footnote{\raggedright A screenshot of the dashboard and the pre- and post-flight procedures are available at \url{https://lis-epfl.github.io/swarm-nxt/flying/}.}}. In addition to executing global and individual flight commands, the dashboard provides comprehensive health monitoring by tracking critical telemetry in real-time, including battery voltage, communication latency, and the operational status of ROS nodes to ensure fleet readiness and identify software failures. It is used in conjunction with Foxglove \cite{foxglove_studio} for high-fidelity visualization of drone poses and planned trajectories.

\section{Experimental Validation} \label{sec:exp_validation}
We validate the SwarmNxt framework with two real-world experiments: a six-drone coordinated flight in a free environment to assess inter-agent collision avoidance, and a four-drone coordinated flight with depth perception in an environment with obstacles to assess obstacle avoidance. \rev{Both were performed indoors in an $8\times8\times4$~m motion-capture arena.}
The vision-based experiment was conducted with four drones as the remaining units did not have fully calibrated fisheye camera setups at the time of the experiments.

\begin{figure*}
  \includegraphics[width=\textwidth]{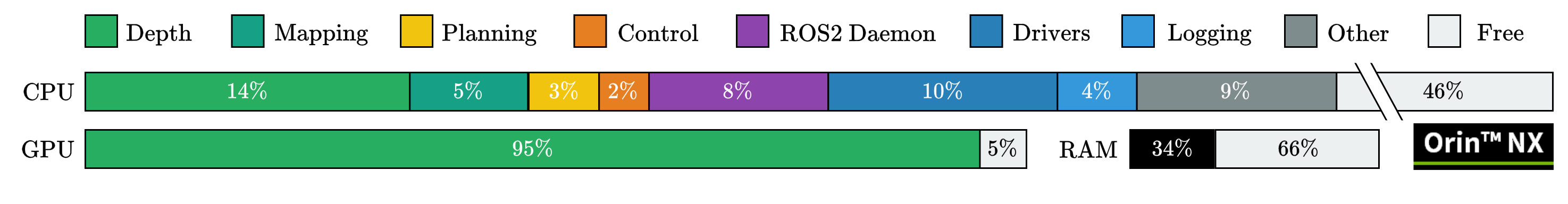}
  \caption{
    Computational resources used by each module of the framework during the experiment with obstacles.
  }
  \label{fig:comp_time}
\end{figure*}

Quantitative metrics are reported for both scenarios over a 2-minute flight duration. \rev{Mapping and planning meet their 100~ms budget even at worst case, and the depth module is free-running, so its $\approx$7~Hz is measured throughput rather than a deadline.
The MPC is the only genuine overrun (40.4~ms worst case against 10~ms); PX4 holds the last setpoint until a new one arrives, and the effect is bounded by the 0.301~m maximum tracking error that the 0.45~m planner safety radius covers.}
The swarm achieves a mean tracking error of approximately 0.075~m and maintains safe inter-agent distances throughout all flights (Table~\ref{tab:swarm_metrics_exact}). When depth estimation is enabled, the autonomy stack utilizes 95\% of the GPU and 54\% of the CPU, leaving sufficient headroom for additional modules on the CPU (Figure~\ref{fig:comp_time}).

\begin{table*}[t]
\centering
\caption{Swarm-wide Computational Performance (SD: Standard Deviation)}
\label{tab:comp_performance}
\small 
\setlength{\tabcolsep}{4.5pt} 
\renewcommand{\arraystretch}{1.3}
\begin{tabular}{l cc ccc ccc} 
\toprule
& \multicolumn{2}{c}{\textbf{Operational Profile}} & \multicolumn{3}{c}{\textbf{4 Drones (with Vision)}} & \multicolumn{3}{c}{\textbf{6 Drones (without Vision)}} \\
\cmidrule(lr){2-3} \cmidrule(lr){4-6} \cmidrule(lr){7-9}
\textbf{Pipeline Stage} & \textbf{Freq. (Hz)} & \textbf{BW (Mbps)} & \textbf{Mean (ms)} & \textbf{SD (ms)} & \textbf{Max (ms)} & \textbf{Mean (ms)} & \textbf{SD (ms)} & \textbf{Max (ms)} \\
\midrule
Depth (S2M2)           & $\approx$7    & 2.551        & 148 & 9.45 & 246 & --- & ---  & --- \\
Mapping (HDSM)         & 10            & 2.051       & 31.8  & 6.98 & 85.8  & 5.17   & 1.28 & 62.6 \\
Motion Planning (HDSM) & 10            & 0.013       & 26.2  & 17.4 & 86.8  & 14.1   &  8.06 & 85.8 \\
Adaptive MPC           & 100           & 0.004       & 1.13   & 1.21 & 40.4  & 0.97   & 1.32 & 39.6 \\
\bottomrule
\addlinespace[0.5em]
\end{tabular}
\end{table*}

\begin{table*}[t]
\centering
\caption{Swarm Tracking Error, Velocity, Jerk, Communication Latency, and Safety Metrics (SD: Standard Deviation)}
\label{tab:swarm_metrics_exact}
\small 
\setlength{\tabcolsep}{3.5pt}
\renewcommand{\arraystretch}{1.3}
\begin{tabular}{l ccc ccc ccc ccc c} 
\toprule
& \multicolumn{3}{c}{\textbf{Tracking Error (m)}} & \multicolumn{3}{c}{\textbf{Velocity (m/s)}} & \multicolumn{3}{c}{\textbf{Jerk (m/s\textsuperscript{3})}} & \multicolumn{3}{c}{\textbf{Com. Latency (ms)}} & \textbf{Safety} \\
\cmidrule(lr){2-4} \cmidrule(lr){5-7} \cmidrule(lr){8-10} \cmidrule(lr){11-13} \cmidrule(lr){14-14}
\textbf{Scenario} & \textbf{Mean} & \textbf{SD} & \textbf{Max} & \textbf{Mean} & \textbf{SD} & \textbf{Max} & \textbf{Mean} & \textbf{SD} & \textbf{Max} & \textbf{Mean} & \textbf{SD} & \textbf{Max} & \textbf{Min. Dist. (m)} \\
\midrule
4 Drones (with vision)    & 0.075 & 0.046 & 0.301 & 1.88 & 1.37 & 4.84 & 13.0 & 8.81  & 30.1 & 4.91 & 12.8 & 41.6 & 0.668 \\
6 Drones (without vision) & 0.077 & 0.048 & 0.315 & 1.93 & 1.32 & 4.99 & 12.9  & 8.74 & 31.4 & 8.53 & 4.51 & 51.7 & 0.652 \\
\bottomrule
\addlinespace[0.5em]
\end{tabular}
\end{table*}

\subsection{Autonomy Stack Configuration}
Identical autonomy stack parameters for the controller, planner, and mapper were used across both experiments. In the free-environment scenario, depth estimation was disabled, as no obstacle avoidance was required. 

The planner safety radius was set to 0.45\,m to account for the worst-case tracking error observed from the MPC. The planning horizon consisted of 12 steps with a temporal resolution of 0.1\,s per step, providing a balance between foresight and computational efficiency. Based on the OmniNxt platform's physical limitations, the following constraints were enforced across all axes: maximum jerk of 20\,m/s\textsuperscript{3}, maximum acceleration of 10\,m/s\textsuperscript{2}, and maximum velocity of 10\,m/s.

\begin{figure*}
\centering
  \includegraphics[width=1.0\textwidth, trim={0cm 0cm 0 0.cm}, clip]{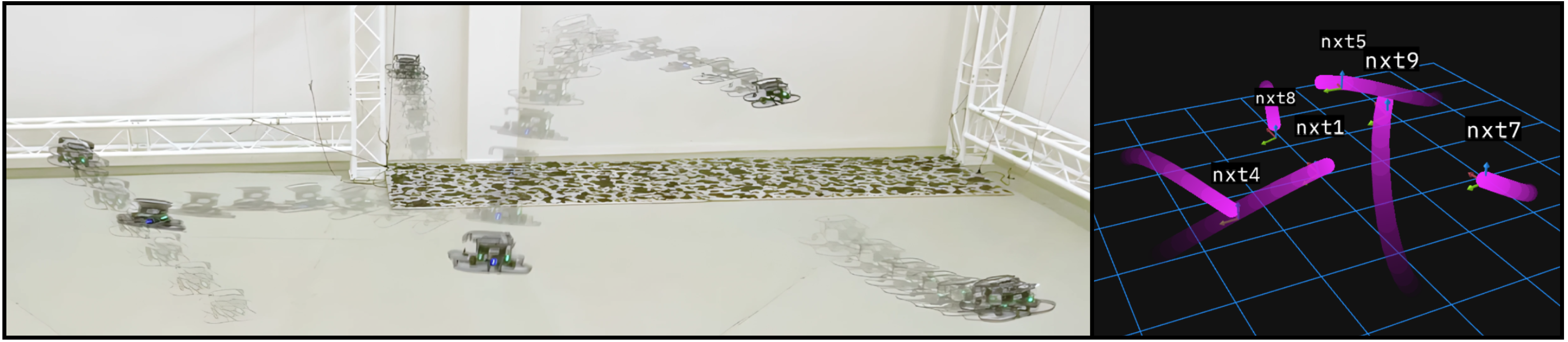}
  \caption{
    The trajectories of the drones during the swap in a free environment (left image), and the visualization of Foxglove during the swap (right image).
  }
  \label{fig:experiment_1}
\end{figure*}

\begin{figure*}
\centering
  \includegraphics[width=1.0\textwidth, trim={0cm 0cm 0 0.cm}, clip]{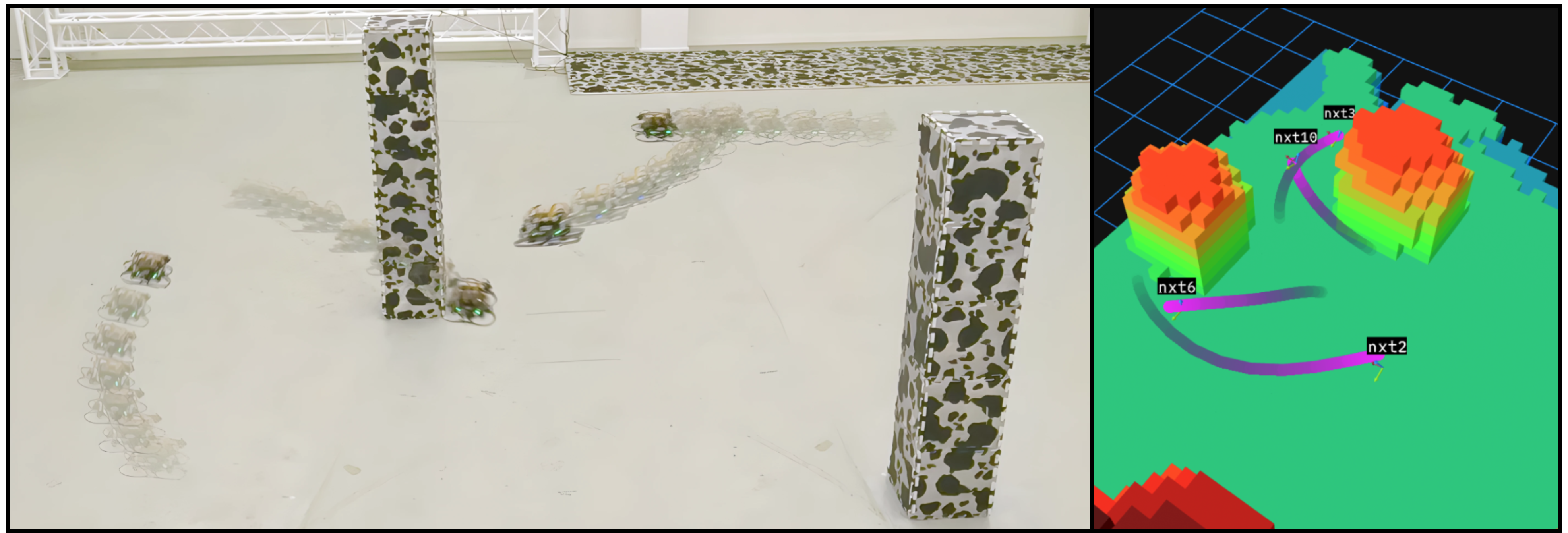}
  \caption{
    The trajectories of the drones while roaming through the obstacles (left image), and the visualization of the occupancy grid of a drone and the trajectories of all drones (right image).
  }
  \label{fig:experiment_2}
\end{figure*}

\subsection{Free Environment}
In the first experiment, we deployed a six-drone swarm in an open environment to validate the platform's ability to handle inter-agent collision avoidance (Figure \ref{fig:experiment_1}). The experiment consisted of two flight phases. First, the drones were initialized in a circle and commanded to swap positions. During the maneuver, they converged toward the center, generating strong aerodynamic disturbances and creating a challenging test for the framework's collision-avoidance capabilities. Then, a continuous roaming experiment was conducted by repeatedly assigning random goal positions to each drone
along the circumference of a circle centered at $(0, 0, 1.5)$\,m with a radius of 2.8\,m.
Once a drone reached its assigned goal within a tolerance of 0.3\,m, a new target position on the circle
was generated.

No collisions were recorded during the cumulative flight time of approximately 2 hours despite communication delay (Table \ref{tab:swarm_metrics_exact}) and packet loss (0.2\%). This is due to the robustness of the planner to these conditions\rev{~\cite{toumiehHighSpeedPlanning}}.

\subsection{Environment with Obstacles}
In the second experiment, we evaluated obstacle avoidance performance using onboard vision-based depth
estimation. A four-drone swarm was deployed in an environment containing static obstacles (Figure \ref{fig:experiment_2}).
The experiment followed the same roaming protocol as in the free-environment scenario, with drones
receiving continuously updated random goals along a circular trajectory.

During the cumulative flight time of 30 minutes, no collisions with obstacles or between agents were recorded.
This is due to our conservative tuning of the planning and mapping modules to avoid collisions. If the environment were to be more cluttered, this conservative tuning would need to be relaxed, and should be paired with more accurate depth estimation.

\section{Conclusion} \label{sec:conclusion}
SwarmNxt is an open-source and adaptable platform for aerial swarm research that builds on the open-hardware OmniNxt. By streamlining the deployment lifecycle—from procurement and hardware assembly to final log analysis—this platform eliminates technical hurdles and facilitates reliable and scalable operation of multiple physical drones. While we used the OmniNxt drone as a physical base for our framework, the software orchestration tools and autonomy
stack can be used with only minor changes on other drones that use the NVIDIA Orin series and PX4.

However, the current S2M2-based depth estimation pipeline is limited in resolution for computational efficiency, which makes it difficult to detect small objects at a distance. In addition, the method requires near-exclusive access to the onboard GPU, preventing the simultaneous execution of other perception modules. Furthermore, the system currently relies on external motion capture because current solutions for Collaborative Visual–Inertial Odometry do not yet provide the accuracy and low-latency performance required for agile swarm flight \cite{xu2024d}.

The work presented here will enable research in those and other challenging areas towards fully autonomous and outdoor deployment of fast, vision-based, and agile drone swarms. For example, this may include robust depth estimation\rev{,} lightweight decentralized relative localization\rev{, multi-agent exploration of unknown environments, and decentralized task allocation}.


\bibliographystyle{IEEEtran}
\bibliography{IEEEabrv,bibliography}

\end{document}